# AI-Augmented LLMs Achieve Therapist-Level Responses in Motivational Interviewing

Yinghui Huang, Yuxuan Jiang, Hui Liu, Yixin Cai, Weiqing Li

*Abstract*—Large language models (LLMs) like GPT-4 show potential for scaling motivational interviewing (MI) in addiction care, but require systematic evaluation of therapeutic capabilities. We present a computational framework assessing user-perceived quality (UPQ) through expected and unexpected MI behaviors. Analyzing human therapist and GPT-4 MI sessions via human-AI collaboration, we developed predictive models integrating deep learning and explainable AI to identify 17 MI-consistent (MICO) and MI-inconsistent (MIIN) behavioral metrics. A customized chain-of-thought prompt improved GPT-4's MI performance, reducing inappropriate advice (Cohen's d = -0.258) while enhancing reflections and empathy. Although GPT-4 remained marginally inferior to therapists overall (OR=1.21), it demonstrated superior advice management capabilities (Cohen's d = -0.478). The model achieved measurable quality improvements through prompt engineering, yet showed limitations in addressing complex emotional nuances. This framework establishes a pathway for optimizing LLM-based therapeutic tools through targeted behavioral metric analysis and human-AI co-evaluation. Findings highlight both the scalability potential and current constraints of LLMs in clinical communication applications.

## I. INTRODUCTION

Globally, behavioral health issues such as substance abuse, chronic disease management, and lifestyle modifications are increasingly becoming major challenges to public health. According to the 2024 World Drug Report by the United Nations, the global prevalence is projected to reach 292 million by 2022, reflecting a 20% increase over ten years. MI, a client-centered counseling approach, has been widely demonstrated to be effective in guiding individuals toward positive behavioral change[1]. The significance of MI lies in its ability to help individuals explore and resolve internal conflicts, thereby eliciting intrinsic motivation and facilitating behavioral change[1].

Recent advancements in LLMs, such as GPT4, have generated significant interest among clinicians, researchers, and the general public for their potential in providing therapeutic support through chatbots[2], [3], [4], [5]. This growing interest is particularly evident in the context of MI, where LLMs are being explored as tools to enhance therapeutic interventions[6]. This growing interest is driven by the stark reality that hundreds of millions of people worldwide lack access to necessary behavioral and mental health care, due to factors such as professional shortages, lack of insurance, high costs, and stigma associated with treatment[7], [8], [9], [10]. These barriers not only degrade individuals' quality of life but also place a significant strain on social and economic development[11].

To address the growing gap in mental health services, individuals and research teams have increasingly turned to LLMs to enhance interactions between therapists and clients[12], [13], [14], [15], [16], [17], [18]. LLM-based self-help interventions in mental health can assist individuals in acquiring therapeutic skills, thereby improving psychological well-being during crisis situations[18]. Concurrently, therapeutic chatbots such as Woebot, Wysa, and Sonia are being developed to provide scalable, personalized support[18], [17]. Notably, many individuals are already using general-purpose LLMs like ChatGPT for therapeutic purposes, even though these models were not specifically designed for such purposes[19], [20], [21].

Both general-purpose and mental health-specific LLMs face rigorous scrutiny due to their potential to generate biased, harmful, or inaccurate information. Mental health professionals have raised significant concerns about the premature integration of LLMs into therapeutic settings without comprehensive evaluation[22], [23], [24]. These concerns are primarily driven by the risk that LLMs may compromise the effectiveness of therapeutic interventions, particularly in the context of MI, where nuanced and empathetic communication is crucial. Furthermore, the use of unverified natural language processing

Yinghui Huang, Yuxuan Jiang and Yixin Cai are with the Research Institute of Digital Governance and Management Decision Innovation, Wuhan University of Technology, Wuhan, 430070, China, and also with School of Management, Wuhan University of Technology, Wuhan, 430070, China. E-mail: huangyh@whut.edu.cn, 338009@whut.edu.cn, 337982@whut.edu.cn.

Hui Liu is with the Key Laboratory of Adolescent Cyberpsychology and Behavior (Central China Normal University), Ministry of Education, Wuhan 430079, China. E-mail: hlccnu@mails.ccnu.edu.cn.

Weiqing Li is with the School of Economics and Management, Hubei University of Technology, Wuhan, 430068, China. E-mail: liwq@mails.ccnu.edu.cn.

Xiangen Hu is with the Learning Sciences and Technologies, Hong Kong Polytechnic University, Hong Kong, 100872, China. E-mail: xiangen.hu@polyu.edu.hk.

Corresponding authors: Hui Liu and Yixin Cai.

This research was supported by funding from the National Natural Science Foundation of China (Award Number: 72204095, 72304090), the Humanities and Social Science Young Scientist Program sponsored by the Ministry of Education of the People's Republic of China (Award Number: 22YJC880022), National Key R&D. Program of China (Grant No.2023YFE0197900 and No.2023YFE0208300), and the China National Center for Mental Health and Prevention, China Education Development Foundation, Ministry of Education Student Service and Quality Development Center (Award Number: XS24A010). None of the funders had any involvement in carrying out this research.

(NLP) interventions with vulnerable populations could introduce substantial risks and ethical challenges, including breaches of confidentiality, the exacerbation of mental health conditions, and the spread of misinformation. As such, it is essential to conduct thorough risk assessments and implement robust mitigation strategies to ensure the safe and ethical integration of LLMs into mental health interventions. This approach is critical for preserving the integrity of therapeutic outcomes and safeguarding the well-being of individuals receiving support.

Established interventions, such as MI, are widely recognized as evidence-based practices for addressing a range of mental health issues, even when delivered through computer-assisted modalities, due to extensive research and rigorous evaluation[1], [25], [26]. While high-quality general-purpose LLMs may have the potential to match the effectiveness of MI delivered by human experts, there is still a lack of sufficient empirical evidence validating the efficacy, safety, and feasibility of using general-purpose LLMs in therapeutic settings, particularly when compared to human therapist. This gap highlights the urgent need to evaluate the feasibility, effectiveness, and safety of LLM-based interventions within the MI framework, in order to accurately assess their potential benefits for patient welfare. Accordingly, this study proposes Research Question 1 (RQ1): To what extent can general-purpose LLMs replicate the performance of human experts in Motivational Interviewing?

In the evaluation of LLMs, human assessments are effective at capturing system performance from the user's perspective, but they are often limited by subjectivity, high costs, inefficiency, and inconsistency[27]. In contrast, automated evaluation methods, while less adaptable, can process large datasets quickly and are praised for their efficiency and consistency[28]. To overcome these limitations, this study introduces a user-centered metric based on UPQ to quantitatively assess LLM performance in MI. Additionally, the study proposes a predictive modeling approach that uses machine learning and deep learning methods to simulate the computational assessment of user-LLM interactions. This dual approach seeks to balance objectivity and efficiency in the evaluation process, while mitigating ethical risks and ensuring that the integration of LLMs into MI practices preserves the integrity and effectiveness of therapeutic interventions.

Understanding the behavior of general-purpose LLMs in high-risk environments, such as mental health, is crucial due to their widespread accessibility and the potential for large-scale adverse outcomes that could severely impact vulnerable clients. These adverse behaviors include clinically inappropriate actions, such as offering solutions before adequately reflecting on a client's emotions and experiences—behaviors that are contraindicated in clinical practice[29]. In the absence of a tailored computational evaluation framework for verbal behaviors, detecting the presence or frequency of such harmful actions remains challenging. Existing research mainly focuses on specific therapeutic modalities[30] and utilizes frameworks like the Cognitive Therapy Rating Scale (CTRS), which were originally designed for the manual assessment of human therapists[31]. However, these frameworks are often complex and not easily adaptable to the automation and large-scale evaluation needs of LLMs. Currently, efforts to translate these established guidelines into metrics suitable for automated and large-scale assessments are limited. This gap underscores the need for specialized computational evaluation frameworks capable of monitoring and mitigating clinically inappropriate behaviors in LLMs, ensuring their safe and ethical deployment in mental health interventions.

In the evaluation of LLMs, intrinsic metrics focus on the quality of the outputs generated by the models—such as coherence, consistency, and relevance—and are essential for understanding verbal behaviors and improving performance across various domains[32], [17]. However, these intrinsic metrics are constrained by their lack of flexibility and limited adaptability to different domains, which can lead to evaluation biases and challenges when addressing novel or unforeseen issues[33]. As a result, intrinsic metrics may not be fully effective for assessing LLM's verbal behaviors specific to MI. In contrast, user-centered extrinsic metrics, such as UPQ, prioritize the impact on users, offering a complementary perspective that is crucial for comprehensive evaluation in high-stakes environments like mental health interventions.

Therefore, this study proposes Research Question 2 (RQ2): How can intrinsic metrics related to UPQ be identified and integrated to develop a framework for evaluating the verbal behaviors of general-purpose LLMs in MI? To address this, we employ explainable machine learning methods to introduce a computational evaluation framework that combines UPQ with relevant intrinsic metrics. This framework incorporates both intrinsic metrics—focusing on desirable and undesirable MI-inconsistent and MI-consistent behaviors—and external metrics—focusing on user impact and perception. Additionally, it elucidates the relationships between these metrics. By integrating these components, the proposed framework aims to enhance the transparency of LLMs within therapeutic processes and foster trust between human users and AI systems[34]. This comprehensive approach ensures that the evaluation of LLMs in MI not only assesses the quality of their language generation but also considers the broader implications for user experience and therapeutic effectiveness.

Identifying both desirable and undesirable behaviors is crucial for understanding the potential benefits and risks associated with LLMs, thereby guiding their adoption, iterative development, and ongoing monitoring. Therefore, it remains to be evaluated whether this framework, particularly the intrinsic metrics related to UPQ, can provide valuable insights to enhance the capabilities of general-purpose LLMs. Given the critical role of prompt engineering in improving LLM performance for specific tasks, this study proposes Research Question 3 (RQ3): Can customized prompts enable general-purpose LLMs to achieve performance comparable to those of professional human experts in MI? Since prompt engineering helps direct LLMs to generate more accurate responses by crafting prompts that better align with user intent and AI comprehension, we integrates insights from the evaluation into customized prompts, reassess the prompted LLM's capabilities to validate the effectiveness of the framework in enhancing LLM performance. This iterative process aims to ensure that LLMs not only meet the technical standards required for effective MI but also adhere to the therapeutic principles necessary for fostering meaningful and supportive interactions with clients.

This study presents an innovative integrated modeling approach for human–AI collaboration in evaluating LLMs[35]. Leveraging a dataset of MI conversations between human and GPT4, the proposed methodology unfolds across three primary steps. First, Predictive Modeling employs machine learning and deep learning techniques to develop a framework for evaluating the UPQ of LLMs in MI contexts. Second, Explainable Modeling identifies key intrinsic metrics that influence UPQ, enabling a detailed assessment of LLM verbal behaviors throughout the MI process. Third, Customized Prompt Development focuses on creating tailored prompts for MI tasks, assessing the efficacy of the evaluation framework in enhancing LLM performance. Additionally, this study conducts a comprehensive comparison between LLMs and human experts across all three steps, offering a thorough analysis of the LLMs' strengths and limitations. The paper is organized as follows: Section 2 reviews relevant literature, including the research background, the principles of MI, and current applications of LLMs in MI. Section 3 outlines the research methodology, covering data collection, framework development, and analytical methods. Section 4 presents the experimental results, while Section 5 discusses the findings, highlights study limitations, and suggests directions for future research. Finally, Section 6 concludes by emphasizing the primary contributions of this study to the fields of MI, mental health, and LLMs.

The contributions of this work are delineated as follows:

(1) This study empirically demonstrated that LLMs achieve performance levels comparable to human therapists in core MI skills, resolving a decades-old limitation in AI's ability to simulate professional psychotherapeutic interactions.

(2) We developed a theory-grounded computational framework based on Miller and Rollnick's MI principles, enabling the first systematic quantification and comparison of verbal behavior patterns between LLMs and human therapists, while providing a reproducible, transparent methodology for evaluating AI agents in psychotherapy contexts.

(3) Through analysis of psychotherapy dialogues, this work revealed LLMs' emergent quasi-social interaction capabilities, thereby expanding the theoretical scope of machine behavior research, and offering empirical evidence for understanding human-AI collaboration mechanisms in high-stakes scenarios.

## II. LITERATURE REVIEW

### A. MI and MI Quality

MI is a client-centered, collaborative counseling approach aimed at enhancing intrinsic motivation and facilitating behavioral change[36]. This approach emphasizes trust-building, addressing ambivalence, and co-developing actionable change plans[37]. Therapist behaviors in MI are classified into MICO, MIIN and other types[38]. MICO behaviors, such as reflective listening and open-ended questioning, help clarify change intentions and enhance intervention efficacy. In contrast, MIIN behaviors, like confrontation and unsolicited advice, may trigger resistance. Other behaviors should be employed contextually to align with intervention goals[39]. Therefore, a comprehensive analysis of therapist behaviors and their impact on MI effectiveness is essential for optimizing therapeutic process and outcomes. MI not only helps individuals resolve behavioral discrepancies but also fosters positive change through collaboration and empathy, making it widely applied in psychotherapy and behavioral interventions[1].

To guarantee the MI quality and guide therapist development, a reliable MI competence assessment tool is crucial[40]. This tool must meet key criteria: inter-rater reliability, minimized evaluator bias[41], and cross-cultural consistency[42]. MI assessment metrics can be categorized into client-reported outcomes and therapist fidelity measures. Client-reported, referring to client behavioral changes such as decreased smoking frequency[43] or alcohol consumption[44], are often evaluated with tools like the Client Evaluation of Motivational Interviewing (CEMI) scale[45]. Therapist fidelity, measured by adherence to MI principles, is assessed with tools like the Motivational Interviewing Treatment Integrity (MITI) scale[41] and the Motivational Interviewing Skills Code (MISC)[46]. While client-reported outcomes directly indicate MI effectiveness, the absence of fidelity assessment may skew results as failing to account for whether the MI principles were effectively implemented[42]. Combining fidelity assessment is crucial for more accurate evaluations of MI competence[47]. Consequently, this paper proposes an integrated approach that merges client-reported outcomes (UPQ) with therapist fidelity (MI strategies and linguistic cues) to offer a more comprehensive and precise evaluation of MI effectiveness, providing a more robust measure of MI quality and a solid foundation for ongoing therapist development.

MI assessment methods can be broadly categorized into manual and automated approaches. Manual evaluations typically rely on standardized tools such as the MITI and MISC[48]. Artificial intelligence (AI) has significantly impacted various industries[49], with machine learning technologies widely applied in finance[50], healthcare[51], and education[52]. In psychological counseling, these technologies are increasingly utilized to automate the evaluation process. For instance, the ReadMI system uses NLP to analyze and assess MI dialogues in real-time[53]. Compared to labor-intensive and time-consuming manual evaluation[54], automated assessment provides a more efficient and scalable solution, facilitating broader adoption and immediate feedback.

### B. Evaluation of Large Language Models in Mental Health Applications

LLMs, based on the Transformer architecture[55], are trained on vast corpora to develop advanced language comprehension and generation[2]. These models can perform a wide range of tasks in a zero-shot manner, without task-specific data[56]. LLMs excel in natural language generation, semantic analysis, and cross-domain applications (e.g., healthcare and education), marking significant advancements in AI technology[57], [58].

LLMs excel at inferring context and generating coherent outputs aligned with prompts. Their integration into mental health is an emerging and promising research area[59]. Studies suggest LLM-powered chatbots can provide on-demand mental health support, eliminate human biases, enhance social confidence, and encourage self-reflection[60]. Furthermore, research has shown that users may feel more at ease sharing personal thoughts with chatbots rather than human therapists[61]. Consequently, LLM-powered chatbots may facilitate expression, assist in psychological issue identification, and support problem-solving.

Despite the promising potential of LLMs in mental health applications, their use involves risks[62]. First, LLMs demonstrate notable inconsistencies when addressing mental health concerns such as anxiety and depression, thereby demonstrating relatively low reliability in these areas[63]. Second, LLMs may generate harmful or inaccurate information[64], [65], posing substantial risks to vulnerable individuals. Even harmful outputs can be misinterpreted as helpful, leading to phenomena like "pathological altruism"[66], and raising ethical concerns about automated decision-making[22]. For instance, weight loss advice given to someone with an eating disorder may worsen their condition, while the same advice might not have such effects in others. Further studies highlight risks of over-reliance on AI, which may contribute to real-life social withdrawal and heightened loneliness[67]. Additionally, the rise of the big data era raises significant concerns about user privacy and data security[68], underscoring the need for strong ethical guidelines in AI-driven mental health tools.

The rapid development of LLMs has increased the demand for robust and standardized evaluation benchmarks[69]. Several scholars emphasize the importance of evaluation as a core discipline crucial to the success of LLMs and other AI systems[70]. LLM evaluation methods can be broadly classified into automated and manual assessments. Though manual evaluation offers real-world relevance and detailed insights[70], it demands substantial financial resources and human labor. Moreover, it suffers from limited reproducibility, vulnerability to evaluator bias, and constraints in assessing LLMs' full potential[71]. Additionally, cultural differences can lead to substantial variability in manual evaluation outcomes[72]. In contrast, automated evaluation offers a more standardized, efficient and consistent process, making it a preferred methodology in practice[70]. Consequently, this paper leverages machine learning and deep learning algorithms to implement automated evaluation, thereby optimizing efficiency and effectiveness.

For LLM's MI performance evaluation, metrics can be categorized into intrinsic and extrinsic types[73]. Intrinsic metrics focus on internal attributes and performance of the model's output, independent of downstream task impact, including factual accuracy, relevance, coherence, and informativeness[74]. In contrast, extrinsic metrics measure the impact of LLM's output on downstream tasks, with helpfulness and effectiveness serving as key evaluation criteria[74]. Intrinsic metrics excel at the linguistic precision and the use of MI strategies, highlighting areas for technical refinement. Extrinsic metrics focus on intervention outcomes, such as promoting behavior change and emotional improvement, thereby highlighting the model's real-world applicability. By integrating both intrinsic and extrinsic metrics, a comprehensive evaluation can be achieved, providing a scientific foundation for enhancing MI quality.

### C. Understanding and Measuring Mental Health Using Digital Language

Language serves as a mirror of attentional focus, revealing underlying emotional states, cognitive processes, and behavioral patterns. Computational linguistics tools facilitate the extraction of nuanced insights into the determinants, correlations, and outcomes of mental health from linguistic data, effectively addressing the limitations of self-report questionnaires and small-scale studies. Empirical studies have demonstrated the capacity of natural language to represent psychological constructs like depression, particularly within multi-method validation frameworks[75]. For instance, language indicative of loneliness (e.g., "alone") is a stronger predictor of depressive mood than simply mentioning "sadness," underscoring the pivotal role of social isolation in depression[76]. Additionally, cognitive manifestations of depression, such as diminished self-worth, often emerge through interrogative terms like 'why' and vague expressions like 'apparently,' indicating how uncertainty exacerbates depression[77]. Furthermore, the intersection of digital language and mental health opens avenues for the development of advanced assessment systems. Research indicates that analyzing digital language patterns can predict depression diagnoses[77], facilitating "early warning systems" that detect pathological states and predict behaviors[78]. These systems have been used to monitor various conditions, including alcohol dependence in trauma patients[79], suicidal ideation[80], adverse childhood experiences in Veterans Affairs patients[81], smoking behavior[66], and post-discharge mood fluctuations[82]. Moreover, predictive language patterns related to mental health, such as expressions of sadness, loneliness, physical discomfort, and hostility, offer clinicians real-time "symptom monitoring dashboards," enabling earlier and more targeted interventions. For example, an increase in loneliness or rumination can signal the need for therapeutic intervention, aligning with MI techniques to foster patient engagement and behavioral change. Upon validation, predictive models based on digital language can be directly applied to assess the mental health status of populations, tracking temporal trends in psychological well-being[75]. Overall, the advancement of non-invasive group measurement methods via digital language offers a robust and scientifically validated approach, serving as a reliable data source for digital epidemiology[83], [84]. This enables real-time, population-wide mental health assessments across all demographic groups, including vulnerable populations, and ensures seamless integration with frameworks for MI and other therapeutic modalities.

NLP[85] is an informatics methodology that extracts structured digital insights from large volumes of unstructured narrative data. Unlike traditional text mining, which focuses on word-level analysis, NLP captures the complexity of unstructured narratives by leveraging semantic relationships and contextual interdependencies[86]. This methodology encompasses syntactic analysis, information extraction, and meaning disambiguation. Recent research has employed NLP to evaluate healthcare quality and safety[87], identify integrated care components in primary healthcare settings[88], and detect alterations in clinical records after disclosure to service users[89]. Within mental health and MI, NLP offers powerful tools for analyzing therapeutic dialogues and patient narratives. By utilizing NLP techniques, researchers can identify linguistic patterns that reflect emotional states, cognitive processes, and behavioral intentions, thus enhancing the understanding of contributors to mental health outcomes and optimizing intervention effectiveness. For instance, NLP can assess MI quality by examining the alignment between client statements and therapeutic strategies, providing real-time feedback to clinicians. Additionally, NLP-driven analysis of patient narratives can reveal underlying themes and sentiments that inform personalized treatment plans, contributing to more targeted and effective mental healthcare. Overall, integrating NLP in mental health and MI frameworks demonstrates the transformative potential of computational linguistics in converting qualitative narrative data into actionable insights, advancing both research and clinical practice.

A dictionary-based approach in NLP utilizes predefined lexicons to identify and extract word categories or phrases from textual data, commonly employed in tasks like part-of-speech tagging, named entity recognition, and sentiment analysis. In this framework, dictionaries consist of predefined lexicons, where words or phrases are mapped to specific categories or labels. One key advantage of dictionary-based methods is their simplicity and ease of implementation, relying on predefined vocabularies without the need for complex algorithms or machine learning models. Linguistic Inquiry and Word Count (LIWC) is a widely used dictionary-based tool, particularly effective in identifying and analyzing emotional states within mental health-related texts[90]. For example, LIWC can quantify the frequency of emotion-related words, cognitive process terms, and social engagement indicators within patient narratives, providing valuable insights into psychological states. Despite their utility, dictionary-based methods have seen limited application in certain areas of mental health interventions. For instance, although these methods have been successfully applied in oncology settings to track changes in clinical records following patient visits[91], their application to clinical records generated by large language model-based chatbots in psychological interventions remains underexplored. This gap highlights an opportunity to leverage dictionary-based NLP techniques to assess changes in dialogues and record-keeping practices post-collaborative sessions with AI-driven conversational agents. In the context of MI, dictionary-based NLP methods are essential for evaluating intervention quality. By systematically classifying and comparing human-AI dialogue content, NLP can provide robust evaluation frameworks for assessing the alignment between therapeutic techniques and client responses. For instance, analyzing the frequency and context of motivational statements, reflections, and affirmations within dialogues can yield insights into intervention effectiveness and the strength of the therapeutic alliance. Overall, research on NLP-driven analysis of mental health intervention quality within dialogue systems remains nascent. However, the integration of dictionary-based approaches can support the development of comprehensive evaluation tools, enhancing the understanding of how AI-mediated interactions influence dialogue quality and therapeutic outcomes. This integration not only facilitates the systematic assessment of intervention fidelity and effectiveness but also contributes to the advancement of personalized and adaptive mental health care.

## III. METHODOLOGY

This study adopted an integrative modeling approach combined with the REFRESH (Responsible and Efficient Feature Reselection guided by SHAP values) methodology to improve predictive performance and model interpretability through dynamic feature selection and iterative updates. The integrative modeling approach strategically combined predictive modeling—focused on outcome forecasting—with explanatory modeling, which sought to elucidate the underlying mechanisms driving observed phenomena. This synthesis ensured both high predictive precision and interpretability. The REFRESH framework optimized performance metrics by dynamically reassessing and reselecting features based on SHAP (SHapley Additive exPlanations) values, which provided a principled way to quantify feature importance. This hybrid approach enhanced model accuracy, fairness, and robustness, while simultaneously maintaining transparency and reliability, thereby establishing a new paradigm for responsible AI applications across diverse contexts[92].

The research utilized publicly available MI demonstration videos from YouTube (Section 3.1), applying both predictive and explanatory modeling within the REFRESH framework to assess GPT-4's performance in MI tasks. In the predictive modeling phase (Section 3.2), two distinct feature engineering methods were employed: LIWC analysis to extract therapist's psychological linguistic features, and pretrained word embedding models (BERT and RoBERTa) to generate dense vector representations of textual data. These feature sets were independently fed into machine learning and deep learning algorithms for training and evaluation, with the goal of identifying the optimal performing model. In the explanatory modeling phase (Section 3.3), key intrinsic metrics influencing UPQ were identified and a computational evaluation framework was developed. The experimental validation phase (Section 3.4) applied customized prompt engineering, informed by the computational evaluation framework, to optimize GPT-4's performance in MI tasks. Finally, GPT-4's performance was assessed based on intrinsic and extrinsic metrics, providing strong empirical evidence for the critical role of the computational framework in enhancing GPT-4's performance in MI contexts. Figure 1 shows a schematic diagram of the whole process.

### A. Data Collection

The dataset employed in this study was curated by a research team led by Zixiu Wu, based on publicly available MI demonstration videos that illustrate typical therapist-client interactions. The team transcribed these videos with high precision, and domain experts specializing in MI meticulously annotated key elements to ensure the dataset's validity, comprehensiveness, and reliability. The dataset captured therapist behaviors (such as reflective listening and questioning) as well as the types of client responses (such as change talk and sustain talk), providing a robust foundation for leveraging machine learning algorithms in the modeling of MI dialogues.

The dataset originally contained only 133 video dialogues and exhibited significant class imbalance, with a disproportionate number of "high" instances in the UPQ category. To address this, this study employed GPT-4 to expand the dataset, resulting in an augmented dataset with 598 video dialogues. This approach effectively mitigated the class imbalance and enhanced the model's generalization and robustness by generating diverse, high-quality text samples, demonstrating the utility of LLMs in NLP data augmentation. The dataset fielded and their corresponding definitions are as follows and the descriptive analysis of the dataset is shown in Table I:

(1) UPQ: Classified as either "high" or "low", used to assess the therapist's MI performance.

(2) Therapist Intervention Type: Includes categories such as "information", "advice", "negotiation" and "options".

(3) Therapist Reflective Listening Type: Divided into simple reflection and complex reflection.

(4) Therapist Question Type: Divided into open-ended and closed-ended questions.

TABLE I
THE DESCRIPTIVE ANALYSIS OF THE DATASET

| Statics | Utterance Rounds | Utterance Length | | |
|---|---|---|---|---|
| | | Overall | Therapist | Client |
| Min | 6 | 2 | 3 | 2 |
| Max | 598 | 1212 | 1034 | 1212 |
| Mean | 25.28 | 93.72 | 103.83 | 83.20 |
| Std | 45.78 | 86.92 | 88.63 | 83.82 |

### B. Evaluating GPT4's Capability in MI Through Predictive Modeling

This study developed a predictive model to assess GPT-4's MI performance, with the UPQ as the dependent variable and linguistic cues and MI strategies derived from the therapist's dialogue as predictors. Linguistic cues were extracted using two distinct methods: the LIWC dictionary and pretrained word embedding models (BERT and RoBERTa). Each method was independently evaluated as a predictor to identify the model yielding the highest performance. A range of machine learning and deep learning classification algorithms were employed for training and prediction, including Random Forest (RF), XGBoost, Support Vector Machines (SVM), as well as Recurrent Neural Networks (RNN), Convolutional Neural Networks (CNN), and Long Short-Term Memory (LSTM).

In the preprocessing stage of feature engineering, the Jieba library, paired with a psychology-specific stopword list, was employed to segment dialogue, and eliminate non-essential terms, thereby isolating core information. Subsequently, linguistic cues reflecting the therapist's behavior were extracted using two distinct methodologies: LIWC and pretrained word embedding models. Each method generated a separate feature set for comparison. LIWC proved an effective tool for capturing key psychological dimensions of the therapist's dialogue, including emotional tone, social engagement, and cognitive processes[93]. Language Style Matching (LSM), a pivotal predictor of MI effectiveness, quantified the linguistic alignment between therapist and client. Greater LSM strengthens the therapeutic alliance, thereby enhancing MI outcomes[94]. LSM was computed based on LIWC analysis using (1) and (2) to measure the alignment of nine functional word categories (e.g., pronouns, prepositions), with the mean score representing the overall LSM, where higher values indicate stronger linguistic alignment[94]. Additionally, pretrained models such as BERT and RoBERTa were employed to generate word embeddings, capturing more nuanced emotional and behavioral patterns in the therapist's speech[95]. Finally, MI strategies, such as reflection and negotiation, were integral in enhancing MI effectiveness and fostering behavioral change[96]. The frequency of these strategies within the dialogue was quantified, with the resulting features as well as linguistic cues incorporated into the predictive model.

$$LSM_{prep} = 1 - \frac{prep_1 - prep_2}{prep_1 + prep_2 + 0.0001} \quad (1)$$

$$LSM = avg \begin{pmatrix} LSM_{prep} + LSM_{article} + LSM_{auxverb} \\ + LSM_{adverb} + LSM_{conj} + LSM_{ppron} \\ + LSM_{ipron} + LSM_{negate} \end{pmatrix} \quad (2)$$

The extracted features were subsequently treated as independent variables, with the initial selection conducted using the Recursive Feature Elimination with Cross-Validation (RFECV) method in combination with the XGBoost model. The synergistic application of RFECV and XGBoost for feature selection significantly mitigates feature redundancy while refining the feature set, thereby improving classification accuracy and model robustness. This approach is particularly well-suited for high-dimensional and imbalanced datasets, leading to marked improvements in model performance for practical applications[97]. The selected features were subsequently fed into multiple classification algorithms, with hyperparameter tuning conducted via GridSearchCV to optimize predictive performance. Model evaluation was carried out using standard classification metrics, including accuracy, precision, recall, F1-score, and ROC-AUC. This methodology facilitated the identification of the optimal predictive model for automated assessment of GPT-4's extrinsic metrics in MI tasks.

Finally, the selected predictive model was used to predict the UPQ of GPT-4's responses obtained through the OpenAI API, and the McNemar test was applied to externally compare the MI competence of GPT-4 with that of human therapists.

### C. Evaluating GPT4's Verbal Behaviors in MI Through Explanatory Modeling

During the explanatory modeling phase, SHAP values were employed to derive a comprehensive global interpretation of the optimal model identified in the predictive modeling phase, while concurrently performing feature selection. This methodology enabled the identification of pivotal intrinsic metrics that exert an influence on the extrinsic metric, thereby contributing to the iterative refinement of the model. Building on the global insights, a subsequent local interpretation was carried out, culminating in the establishment of a UPQ-centered computational evaluation framework.

SHAP values, grounded in the Shapley value principle from cooperative game theory, provide a consistent and equitable allocation of feature contributions, thereby making complex black-box models interpretable. This methodology is broadly applicable and instrumental in identifying the underlying factors driving model predictions, thus ensuring the reliability and transparency of interpretative outcomes. SHAP values supported both local and global interpretations, offering insights into the overall model behavior and the specific contributions of key features[98]. Within the REFRESH framework, SHAP values were computed to provide a global interpretation based on the optimal predictive model identified during the predictive modeling phase, thereby allowing for the quantification of the cumulative contribution of each intrinsic metrics and its directional influence on the extrinsic metric. Features were ranked by cumulative contribution and sequentially incorporated into the predictive model in descending order for forward stepwise selection and cross-validation, aiming to identify the top n features that optimize model performance. This methodology improved model performance[99] and enhanced interpretability[100], mitigated overfitting[101], and illuminated the relationships between features and the target variable, thereby ensuring heightened transparency and robustness of the model[102]. Subsequently, a local interpretation of the refined model was performed, leading to the development of a computational evaluation framework, which includes the identification of intrinsic metrics with a significant impact on extrinsic metrics and their corresponding influence directions. Finally, based on the computational evaluation framework, a paired-sample t-test was conducted to internally assess the MI competence of GPT-4.

### D. Evaluating the Framework's Enhancement in GPT's Performance in MI Through Prompt Engineering

The CoT prompting framework, a key technique in prompt engineering, enhances a model's capacity to tackle complex tasks and improves the logical coherence of its outputs by guiding step-by-step reasoning[103]. Based on the computational evaluation framework established in the explanatory modeling phase, a customized CoT-based prompt was developed to optimize GPT-4's performance in MI tasks. The effectiveness of this customized prompt was rigorously evaluated by comparing the extrinsic and intrinsic metrics of the prompted GPT-4 responses against those of human therapists and the baseline GPT-4 model, thereby validating the effectiveness of the computational evaluation framework in enhancing GPT-4's MI capabilities.

### E. Summary

The research framework of this study is illustrated in Fig. 1. The dataset utilized was systematically curated by Zixiu Wu, drawing upon publicly available MI videos from the YouTube platform. The dialogue content in these videos was rigorously annotated by multiple subject matter experts, systematically capturing the therapists' MI strategies as well as other relevant contextual factors.

Secondly, in the predictive modeling phase, feature engineering generated two distinct feature sets: therapists' linguistic cues (including LIWC analysis results and LSM) or word embeddings, and MI strategies. These feature sets were subsequently used as independent variables in various machine learning and deep learning classification algorithms for training. The models were evaluated using standard classification metrics, which allowed for the identification of the highest-performing model. The optimal model was then applied to predict the extrinsic metric of GPT-4's responses, and McNemar's test was conducted to compare the UPQ between GPT-4 and human therapists aimed at evaluating GPT-4's MI performance.

In the explanatory modeling phase, the optimal feature set derived from the predictive modeling phase was fed into the predictive model for a global SHAP analysis, and the features were ranked according to their cumulative contributions. The features were subsequently introduced into the predictive model in descending order for forward stepwise feature selection and cross-validation to refine the feature set and improve model performance. A local SHAP analysis was then conducted on the refined model to establish a computational evaluation framework. Finally, a paired-sample t-test within the computational framework evaluated GPT-4's MI competence.

During the validation phase, the computational evaluation framework was employed to design CoT-based customized prompts to enhance GPT-4's performance. These prompts were then deployed via the OpenAI API to generate GPT-4-prompted responses. McNemar's test and paired-sample t-tests were conducted to compare the extrinsic and intrinsic metrics between human therapists and GPT-4-prompted responses, thereby validating whether the computational framework improved GPT-4's MI performance.

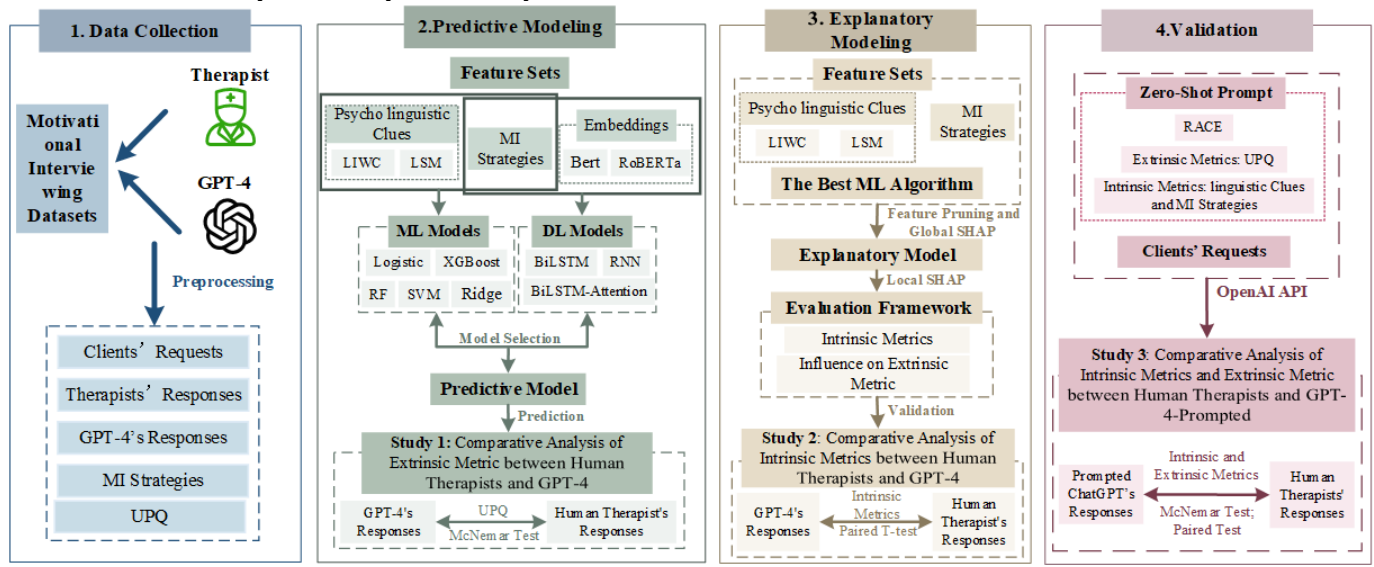

Fig. 1. This figure illustrated the step-by-step workflow of the proposed work. The workflow commenced with data collection and preprocessing. Subsequently, the predictive modeling phase employed two types of feature sets alongside multiple machine-learning and deep-learning models to predict the MI capabilities of GPT-4 and conducted a comparison with human therapists in terms of extrinsic metrics. Next, the explanatory modeling phase established a computational evaluation framework to assess GPT-4's MI capabilities from both intrinsic and extrinsic metrics. Finally, the validation phase generated prompts based on the computational evaluation framework, compared the MI capabilities of GPT-4 and human therapists after targeted prompting, so as to verify the correctness of the computational evaluation framework.

## IV. RESULT

This section will sequentially present the results of predictive modeling and explanatory modeling, the development and evaluation of the Integrative Evaluation Framework, ethical risks and potential biases, and concluding with limitations and future research.

### A. Evaluation of GPT-4's Performance in Motivational Interview Based on UPQ by Predictive Modeling

In this study, multiple classification algorithms were trained and evaluated, with the most effective model selected for the development of a predictive model to assess the UPQ of GPT-4's responses. The UPQ scores of GPT-4's responses were then compared to those of human therapists to evaluate the extent to which GPT-4's MI performance approximates human-level capabilities.

*1) Performance Analysis of the Predictive Model for UPQ*

Following feature engineering, classification models were trained and evaluated based on two distinct sets of independent variables. The first set encompassed therapists' linguistic features extracted from the LIWC tool, LSM between the therapist and client, and the frequency of MI strategies employed by the therapist. The second set consisted of word embeddings for both therapists and clients, generated by pretrained language models (BERT and RoBERTa), as well as the frequency of MI strategies. A range of classification algorithms were utilized to construct predictive models, and the corresponding performance metrics were presented in Table II.

TABLE II
GRAND-AVERAGE PREDICTED PERFORMANCES OF DIFFERENT CLASSIFICATION MODELS (ACCURACY, PRECISION, RECALL, F1-SCORE, AND ROC AUC)

| Features | | Model | Accuracy | Precision | Recall | F1-Score | ROC AUC |
|---|---|---|---|---|---|---|---|
| LIWC | Machine Learning | RF | 0.8940 | 0.9089 | 0.8848 | 0.8958 | 0.9597 |
| | | SVM | 0.9636 | 0.9742 | 0.9575 | 0.9653 | 0.9923 |
| | | Logistic | 0.9525 | 0.9394 | 0.9583 | 0.9483 | 0.9898 |
| | | Ridge | 0.9426 | 0.9351 | 0.9416 | 0.9381 | 0.9870 |
| | | Xgboost | 0.9145 | 0.9299 | 0.9059 | 0.9168 | 0.9704 |
| | Deep Learning | RNN | 0.7650 | 0.8068 | 0.5167 | 0.6238 | 0.8192 |
| | | Bilstm | 0.8158 | 0.8307 | 0.6592 | 0.7302 | 0.8897 |
| | | Bilstm-Attention | 0.7463 | 0.7431 | 0.5456 | 0.6190 | 0.7455 |
| Bert | Machine Learning | RF | 0.8252 | 0.8271 | 0.6257 | 0.7108 | 0.9037 |
| | | SVM | 0.8326 | 0.8329 | 0.6525 | 0.7296 | 0.8557 |
| | | Logistic | 0.8238 | 0.8097 | 0.8281 | 0.8182 | 0.9059 |
| | | Ridge | 0.8297 | 0.8097 | 0.8282 | 0.8182 | 0.9059 |
| | | Xgboost | 0.9153 | 0.8676 | 0.8963 | 0.8807 | 0.9740 |
| | Deep Learning | RNN | 0.8290 | 0.8474 | 0.6273 | 0.7181 | 0.8742 |
| | | Bilstm | 0.8384 | 0.8798 | 0.6263 | 0.7297 | 0.8843 |
| | | Bilstm-Attention | 0.8422 | 0.8887 | 0.6306 | 0.7364 | 0.8760 |
| Roberta | Machine Learning | RF | 0.7849 | 0.7985 | 0.8046 | 0.7972 | 0.8775 |
| | | SVM | 0.7864 | 0.7939 | 0.8151 | 0.8004 | 0.8665 |
| | | Logistic | 0.8238 | 0.8163 | 0.8038 | 0.8081 | 0.9164 |
| | | Ridge | 0.8337 | 0.8207 | 0.8246 | 0.8216 | 0.9255 |
| | | Xgboost | 0.9190 | 0.8675 | 0.9059 | 0.8857 | 0.9755 |
| | Deep Learning | RNN | 0.8403 | 0.8643 | 0.6539 | 0.7413 | 0.8926 |
| | | Bilstm | 0.8309 | 0.8381 | 0.6357 | 0.7196 | 0.8833 |
| | | Bilstm-Attention | 0.8102 | 0.8190 | 0.5772 | 0.6753 | 0.8221 |

The evaluation results revealed that, in general, machine learning models outperformed deep learning models in terms of predictive performance. Specifically, the accuracy of machine learning models ranged from 0.7849 to 0.9636, with ROC AUC values spanning from 0.8557 to 0.9923. In contrast, deep learning models demonstrated relatively weaker performance, with accuracy values between 0.7463 and 0.8422, and ROC AUC values ranging from 0.7455 to 0.8926. Furthermore, machine learning models utilizing LIWC feature set consistently outperformed those based on word embeddings. For example, the SVM model, incorporating LIWC features, achieved the highest accuracy (0.9636) and ROC AUC (0.9923). By contrast, the XGBoost model, which incorporates word embeddings derived from BERT or RoBERTa, exhibited optimal performance within its respective class of models, achieving accuracies of 0.9153 and 0.9190, alongside ROC AUC values of 0.9740 and 0.9755, respectively. Overall, the SVM model with the LIWC feature set emerged as the top performer, demonstrating outstanding classification accuracy and discriminative power. These findings underscored the significant advantage of combining LIWC features with machine learning algorithms for predictive tasks, particularly in evaluating MI performance.

*2) Comparison Analysis of UPQ Between GPT-4 and Human Therapists with Clients*

The predictive model, an SVM classifier based on LIWC features, was utilized to assess the UPQ of responses generated by GPT-4. Given that the UPQ of human therapists was represented as a binary variable, the McNemar test was employed for statistical analysis, alongside visualizations of the UPQ distribution for both groups. The odds ratio (OR) was additionally utilized as an effect size measure to quantify the differences in UPQ between the two groups. As indicated in Fig. 2 and Table III, human therapists achieved a significantly higher UPQ (52.69% vs. 38.45%, $\chi^2$ = 25.886, p < 0.001). The OR value suggested that the probability of high UPQ in the human therapist group was 37% greater than that in the GPT-4 group. These results address RQ1.

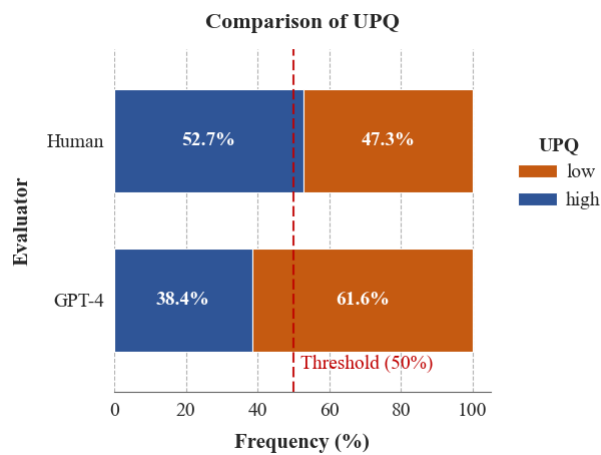

Fig. 2. This figure compares the UPQ scores between human therapists and GPT-4, presenting the frequency percentages of low and high UPQ values.

TABLE III
COMPARISON OF HIGH UPQ PERFORMANCE BETWEEN HUMAN THERAPISTS AND GPT-4 USING MCNEMAR TEST, INCLUDING CHI-SQUARE ($\chi^2$), P-VALUE (P), AND ODDS RATIO (OR)

| Project | N | UPQ | | McNemar Test | | |
|---|---|---|---|---|---|---|
| | | Human High UPQ | GPT-4 High UPQ | $\chi^2$ | P | OR |
| Human therapist vs GPT-4 | 632 | 333(52.69%) | 243(38.45%) | 25.886 | <0.001 | 1.47 |

### B. Evaluating GPT-4's Linguistic Behavior in Motivational Interview via Intrinsic Metrics by Explanatory Modeling

To gain deeper insights into and evaluate GPT-4's MI performance, we developed a computational evaluation framework. Initially, the global SHAP method was employed for sensitivity analysis and local SHAP method for feature pruning, leading to the construction of a more refined predictive model. These results were then integrated with relevant theoretical paradigms to establish a comprehensive, client-centered computational evaluation framework tailored for generative AI. This framework enabled an in-depth quantitative assessment of responses generated by GPT-4 and human therapists.

*1) Extraction of intrinsic metrics: Identifying psychological linguistic cues that influence UPQ through explainable ML*

The initial feature set in this study consisted of 127 variables, which was subsequently reduced to 21 through feature selection with RFECV. In this section, the global and local SHAP methods were utilized for sensitivity analysis and feature selection to enhance the interpretability of the predictive model.

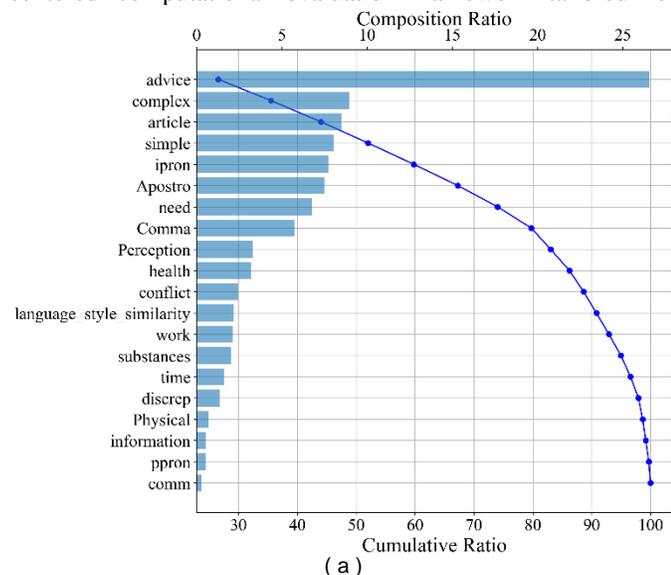
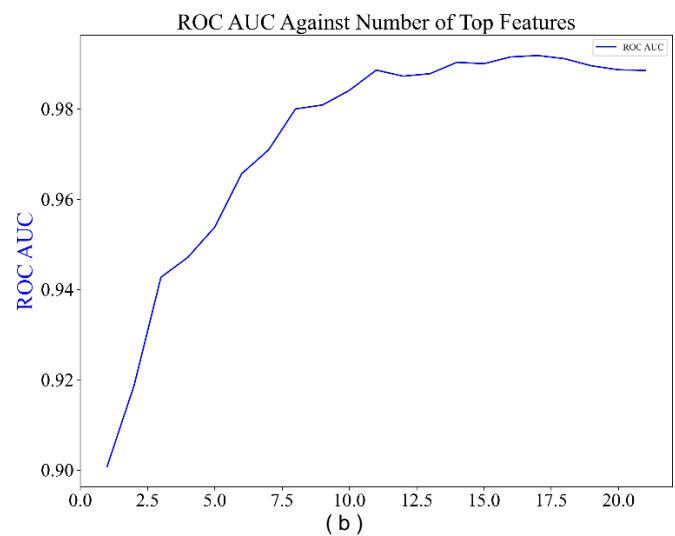

Fig. 3. This figure presents the composition ratio of top predictive features ranked by SHAP value importance (left) and ROC AUC performance against the number of top features (right) for feature selection evaluation.

First, SHAP values were computed to rank features based on their relative importance, as illustrated in Fig. 3 (a). Following this, forward stepwise feature selection approach was performed, where features were incrementally added to the model in descending order of their SHAP importance. Given the class imbalance in the dataset, the performance of the updated model was evaluated using the ROC AUC metric after each feature addition. As illustrated in Fig. 3 (b), the model achieved a local optimum with 17 features, reaching an accuracy of 0.9703 and an ROC AUC of 0.9918, resulting in a more refined and interpretable model. As demonstrated in Table IV, the accuracy of the pruned model slightly outperformed that of the original model (0.9636), accompanied by a modest increase in ROC AUC, rising from 0.9898 to 0.9918 in the pruned model, further underscoring the enhanced performance and efficiency of the pruned model.

Table IV
COMPARISON OF MODEL PERFORMANCE BEFORE AND AFTER FEATURE PRUNING

| Model | Accuracy | ROC AUC |
|---|---|---|
| Original SVM Model | 0.9636 | 0.9898 |
| Pruned SVM Model | 0.9703 | 0.9918 |

As illustrated in Fig. 4, a local interpretability analysis utilizing SHAP values was conducted to provide a more comprehensive understanding of both the magnitude and direction of each feature's influence on UPQ, subsequent to the global SHAP analysis.

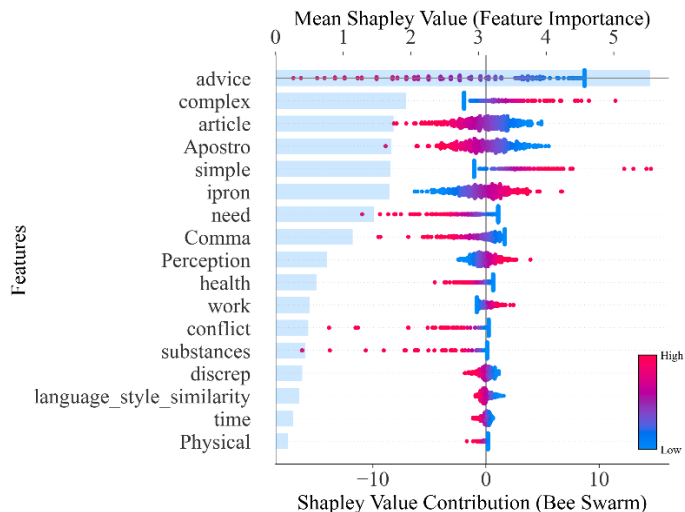

Fig. 4 This figure presents the SHAP-based local interpretability analysis of feature contributions, including feature importance and directional effects.

Table V
THE IMPACT OF INTRINSIC METRICS: IMPORTANCE RANKING AND INFLUENCE ON UPQ

| Ranks of Intrinsic Metrics | Categories of Intrinsic Metric | Intrinsic Metric | Relation Between Features and the UPQ | Importance Scores |
|---|---|---|---|---|
| 1 | Advice (without permission) | advice | Negative | 5.540897 |
| 2 | REC (Complex Reflection) | complex | Positive | 1.930714 |
| 3 | Analytical Thinking | article | Negative | 1.746483 |
| 4 | Professionalism | Apostrophes | Negative | 1.710157 |
| 5 | RES (Simple Reflection) | simple | Positive | 1.700861 |
| 6 | Analytical Thinking | impersonal pronouns | Positive | 1.685375 |
| 7 | Direct | need | Negative | 1.457068 |
| 8 | Empathy | Comma | Negative | 1.14083 |
| 9 | Empathy | Perception | Positive | 0.7562 |
| 10 | Warn | health | Negative | 0.605685 |
| 11 | Empathy | work | Positive | 0.502621 |
| 12 | Confrontation | conflict | Negative | 0.484878 |
| 13 | Warn | substances | Negative | 0.434961 |
| 14 | Raise Concern (without permission) | discrepancy | Negative | 0.39475 |
| 15 | Negotiation | LSM | Negative | 0.35326 |
| 16 | Structure | time | Negative | 0.261476 |
| 17 | Warn | Physical | Negative | 0.182461 |

As presented in Table V, these intrinsic metrics were aligned with the MI dimensions from existing literature, facilitating an exploration of the relationships between the intrinsic and extrinsic metrics. Based on the importance scores, the features were ranked in descending order to establish a UPQ-centered computational evaluation framework for MI.

This analysis delineates how intrinsic metrics shape the MI quality. The "advice" feature in the "Advice (without permission)" category exhibited a strong negative correlation with UPQ, recording the highest importance score (5.540897). In contrast, the "Reflection" category revealed positive correlations with UPQ for both "Complex Reflection" (1.930714) and "Simple Reflection" (1.700861). In the "Analytical Thinking" dimension, "article" showed a negative association with UPQ (1.746483), whereas "impersonal pronouns" were positively linked to UPQ (1.685375). The "Apostrophes" feature under the "Professionalism" category was negatively associated with UPQ (1.710157). Within the "Direct" category, the "need" feature demonstrated a negative relationship with UPQ (1.457068). Regarding the "Empathy" category, both "Perception" and "work" (0.502621) were positively correlated with UPQ (0.7562), while "Comma" (1.14083) exhibited negative associations. The "LSM" in "Negotiation" category showed a negative correlation with the UPQ (0.35326). In the "Warn" category, the features such as "health", "substances", and "Physical" were negatively associated with UPQ, with importance scores of 0.605685, 0.434961, and 0.182461, respectively. Additional negatively correlated features included "conflict" within the "Confrontation" category (0.484878) and "discrepancy" in the "Raise Concern (without permission)" category (0.39475). The "time" feature under the "Structure" category also showed a negative association with UPQ (0.261476). In summary, features with higher importance scores and negative associations, such as "advice", "articles" and "Apostrophes", exerted a significant influence on UPQ, while features positively correlated with UPQ, such as "Complex Reflection" and "Simple Reflection", facilitated an enhancement in UPQ.

*2) Evaluating GPT-4's performance Motivational Interview in by intrinsic metrics*

To examine the impact of intrinsic metrics on UPQ, paired-sample t-tests were employed to compare the deployment of MI strategies and linguistic cues between human therapists and GPT-4, thereby elucidating differences in their performance across these intrinsic metrics. The results are summarized in Table VI.

Table VI
COMPARATIVE ANALYSIS OF GPT-4 AND HUMAN THERAPISTS: PERFORMANCE ON INTRINSIC AND EXTRINSIC METRICS[1]

| Categories | Categories of Intrinsic Metric | Intrinsic Metric | Relation Between Features and the UPQ | Cohen's d (GPT-4 - Human Therapists) | Importance Scores |
|---|---|---|---|---|---|
| MIIN | Advice (without permission) | advice | Negative | -0.234*** | 5.540897 |
| | Direct | need | Negative | 0.151*** | 1.457068 |
| | Warn | Physical | Negative | -0.195*** | 0.182461 |
| | | health | Negative | -0.169*** | 0.605685 |
| | | substances | Negative | -0.125** | 0.434961 |
| | Raise Concern (without permission) | discrepancy | Negative | 0.632*** | 0.39475 |
| | Confrontation | conflict | Negative | 0.068 | 0.484878 |
| MICO | Reflection | RES (simple reflection) | Positive | -0.218*** | 1.700861 |
| | | REC (complex reflection) | Positive | 0.166*** | 1.930714 |
| Other | Empathy | Perception | Positive | 0.396*** | 0.7562 |
| | | Comma | Negative | -0.571*** | 1.14083 |
| | | work | Positive | -0.384*** | 0.502621 |
| | Professionalism | Apostrophes | Negative | 0.782*** | 1.710157 |
| | Negotiation | LSM | Negative | 0.478*** | 0.35326 |
| | Structure | time | Negative | 0.014 | 0.261476 |
| | Analytical Thinking | impersonal pronouns | Positive | -0.060 | 1.685375 |
| | | article | Negative | 0.540*** | 1.746483 |

First, as illustrated in Table VI, although GPT-4's overall performance on intrinsic metrics marginally lagged behind that of human therapists, it exhibited a distinct advantage in the MIIN behaviors. Specifically, GPT-4 outperformed human therapists in "Advice (without permission)" dimension (Cohen's d = –0.234, p < 0.001), and across all subcategories of "Warn", including "Physical" (Cohen's d = –0.195, p < 0.001), "Health" (Cohen's d = –0.169, p < 0.001), and "Substances" (Cohen's d = –0.125, p < 0.01). However, in the MI process, GPT-4's increased expression frequency in the "Direct" dimension (Cohen's d = 0.151, p < 0.001) and the "Raise Concern (without permission)" dimension (Cohen's d = 0.632, p < 0.001) detrimentally influenced UPQ.

Second, GPT-4's performance on MICO behaviors aligns with that of human therapists, with a particular emphasis on the "Reflection" dimension in this study. Notably, GPT-4 employed "Complex Reflection" at a significantly higher frequency compared to human therapists (Cohen's d = 0.166, p < 0.001), which led to a marked improvement in the UPQ. In contrast, GPT-4 showed a significant reduction in the use of "Simple Reflection" (Cohen's d = -0.218, p < 0.001), which negatively impacted the UPQ.

Third, GPT-4's performance in other behaviors generally lagged behind that of human therapists. In the "Empathy" dimension, GPT-4 demonstrated an increase in the use of "Perception" expressions (Cohen's d = 0.396, p < 0.001) and a decrease in the use of "Comma" (Cohen's d = –0.571, p < 0.001). However, GPT-4 displayed a marked reduction in its engagement with "Work" discussions (Cohen's d = –0.384, p < 0.001). In the "Professionalism" dimension, GPT-4 demonstrated a higher frequency of "Apostrophes" (Cohen's d = 0.782, p < 0.001). In the "Negotiation" dimension, GPT-4 showed an increase in LSM (Cohen's d = 0.478, p < 0.001), and in the "Analytical Thinking" dimension, there was a marked increase in the use of "Articles" (Cohen's d = 0.540, p < 0.001), all of which may have hindered the enhancement of UPQ. Furthermore, GPT-4 showed no significant difference in the use of "impersonal pronouns" compared to human therapists (Cohen's d = –0.060, p > 0.05). In the same time, in the "Structure" dimension, GPT-4's use of "Time" (Cohen's d = 0.014, p > 0.05) did not differ significantly from that of human therapists.

---

[1] Asterisks in Table VI denote levels of statistical significance: * for p < 0.05, ** for p < 0.01, and *** for p < 0.001.

From the perspective of feature importance in predicting UPQ, human therapists generally outperformed GPT-4 on most key intrinsic metrics with substantial impacts on UPQ, such as "Simple Reflection", "Apostrophes", "articles", and "need". However, GPT-4 surpassed human therapists only in "Advice" and "Complex Reflection". These results indicate that while GPT-4 exhibited notable strengths in specific intrinsic metrics, its overall performance remained inferior to that of human therapists. Consequently, to improve GPT-4's performance on extrinsic metrics, it is pivotal to prioritize enhancements in its linguistic cue sand MI strategies, given that these foundational metrics constitute the cornerstone for meaningful advancements in both overall assessment outcomes and therapeutic effectiveness. These results address RQ2.

### C. Validation of the Integrative computational Evaluation Framework

To validate the pivotal role of the computational evaluation framework in improving GPT-4's MI performance, this study employed zero-shot prompting strategies informed by prior findings to design the CoT prompt engineering method grounded in the framework. Responses generated by GPT-4 using this customized prompt were systematically compared to those of human therapists.

*1) Developing zero-shot prompts: Utilizing insights from UPQ-centered explanatory modeling to enhance GPT-4's responses*

Table VII
PROMPT FRAMEWORK FOR ENHANCING GPT-4'S UPQ IN MOTIVATIONAL INTERVIEW

| Dimensions of MI | Categories of Intrinsic Metric | Strategies promoting UPQ | Strategies prompting Intrinsic Metrics | | Customized prompts |
|---|---|---|---|---|---|
| | | | Increase Usage | Decrease Usage | |
| MIIN | Advice (without permission) | Seek consent and encourage client-driven solutions to foster collaboration. | - | advice | Avoid advice without permission to boost autonomy, reduce resistance, and strengthen the alliance. |
| | Direct | Reduce the use of directive language to enhance collaboration and autonomy, and decrease resistance. | - | need | Avoid phrases like 'need' or 'must,' and foster collaboration to support client-driven solutions. |
| | Warn | Avoid warning behaviors that focus on risks, as they may increase resistance. | - | physical / health / substance | Foster collaboration and explore the client's motivations, minimizing emphasis on health risks like smoking or drinking. |
| | Raise Concern (Without Permission) | Always seek permission before raising concerns to align with client readiness and respect their autonomy. | - | discrepancy | Avoid excessive focus on discrepancies and always seek permission before addressing them. |
| | Confrontation | Foster a collaborative, non-judgmental environment to encourage open dialogue. | - | conflict | Avoid conflict-related language, as they damage the therapeutic alliance and increase resistance. |
| MICO | Reflection | Use reflective listening to validate feelings, deepen exploration, and strengthen collaboration. | RES (Simple Reflection) / REC (Complex Reflection) | - | Combine simple reflections to rephrase client words and foster rapport with complex reflections to infer deeper meanings and evoke change talk. |
| Other | Empathy | Show high empathy to build trust, reduce resistance, and foster motivation for change. | Perception / work | Comma / - | Show empathy by reflecting emotions, perceptions, and work-related topics. Avoid long pauses (Comma). |
| | Professionalism | Use formal, precise language to build trust and strengthen the alliance. | - | Apostrophes | Avoid casual elements like overused apostrophes that undermine credibility. |
| | Negotiation | Negotiation aligns goals, resolves ambivalence, and fosters commitment to change. | - | LSM | Instead of overemphasizing LSM, focus more on the tasks within the negotiation strategy during the MI process |
| | Structure | Use structure and timelines sparingly to avoid disrupting | - | time | Minimize rigid transitions and time focus to support |

| | | | |
|---|---|---|---|
| | flow or reducing autonomy. | | engagement. |
| Analytical Thinking | Overuse of analytical language can make therapy feel impersonal, weakening empathy and therapeutic alliance. | impersonal pronouns | article | Analytical thinking, marked by high article use and low pronoun use, may overly focus on logical clarity, reducing emotional connection. |

As demonstrated in Table VII, this study aimed to enhance the response quality of GPT-4 in the MI task by designing a customized prompt based on the computational evaluation framework. Leveraging a CoT methodology, the customized prompt was systematically integrated with the original RACE prompt to create a tailored zero-shot prompt. Comprehensive details regarding both the original and customized prompts are provided in supplementary materials.

*2) Comparative analysis of the UPQ between GPT-4-Prompted and Human Therapists in MI*

To validate the effectiveness of the computational evaluation framework in enhancing GPT-4's response performance on the extrinsic metric, a McNemar test was conducted to examine whether significant differences in UPQ existed across three groups of responses, with OR as the effect size. The results, presented in Fig. 5 and Table VIII, revealed a statistically significant improvement in UPQ for GPT-4-Prompted responses compared to standard GPT-4 responses at a significance level of 0.05 ($\alpha = 0.05$) (46.84% vs. 38.45%, $\chi^2 = 0.292$, p = 0.002). However, compared to human therapists, the UPQ of GPT-4 enhanced by customized zero-shot prompts remained marginally lower (46.84% vs. 52.69%, $\chi^2 = 4.966$, p = 0.026). The OR values indicated that while GPT-4-Prompted responses were 21% more likely to achieve high UPQ compared to standard GPT-4 responses, they were 19% less likely to do so than human therapists.

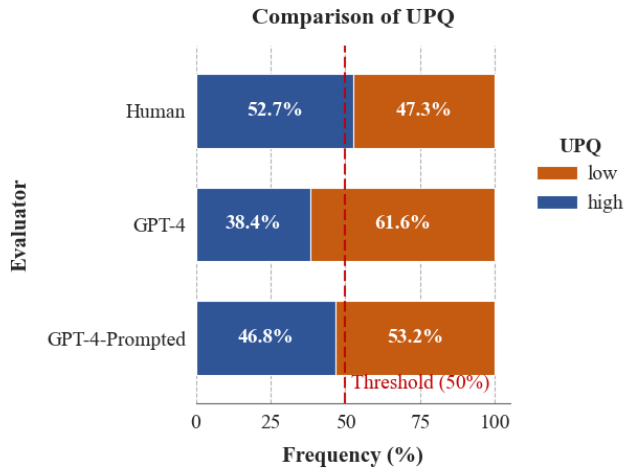

Fig. 5 This figure compares the UPQ scores between human therapists, GPT-4 and GPT-4-Prompted, presenting the frequency percentages of low and high UPQ values.

Table VIII
COMPARISON OF HIGH UPQ PERFORMANCE BETWEEN GPT-4-PRMOPTED AND HUMAN THERAPISTS AS WELL AS GPT-4 USING MCNEMAR TEST, INCLUDING CHI-SQUARE ($\chi^2$), P-VALUE (P), AND ODDS RATIO (OR)

| Project | N | McNemar Test | | |
|---|---|---|---|---|
| | | $\chi^2$ | P | OR |
| GPT-4-Prompted vs Human Therapists | 632 | 4.966 | 0.026 | 0.81 |
| GPT-4-Prompted vs GPT-4 | 632 | 0.292 | 0.002 | 1.21 |

*3) Comparison Analysis in the Intrinsic Metric between GPT-4-Prompted and Human Therapists*

To assess whether customized prompts enhance GPT-4's performance on intrinsic metrics, a paired-sample t-test was conducted to compare the use of intrinsic metrics between GPT-4-Prompted and standard GPT-4 responses, as well as between GPT-4-Prompted responses and those of human therapists in MI. Results are detailed in Table IX.

First, a comparative analysis between GPT-4-Prompted and human therapists revealed that, although GPT-4-Prompted remained less proficient than human therapists in overall intrinsic metrics, it exhibited domain-specific advantages. Specifically, within the MIIN dimension, GPT-4-Prompted demonstrated a substantial reduction in the use of "Advice (without permission)" (-0.478***). Additionally, GPT-4-Prompted reduced the frequency of "Warn" associated with substance-related vocabulary (-0.138***). Next, in the other behaviors dimension, GPT-4-Prompted achieved a marked increase in the use of "Perception" in the Empathy dimension (0.559***).

Second, compared with the baseline GPT-4, the customized prompting methodology demonstrated clear enhancements in intrinsic metrics. Specifically, in the MIIN dimension, a significant reduction in "Advice (without permission)" (-0.258***) usage positively influenced the UPQ. In the MICO dimension, the use of "Simple Reflection" (0.093*) increased. In the other behaviors, GPT-4-Prompted exhibited significant increases in "Perception" (0.146***) and "Work" (0.136***) discussions, reflecting higher levels of empathy. Additionally, a marked reduction in "Apostrophes" (-0.732***) contributed to enhanced professionalism. Furthermore, the decrease in LSM (-0.145***) facilitated improvements in UPQ scores. At the same time, the reduction in the use of "Articles" (-0.147***) effectively lowered the prevalence of "Analytical Thinking".

Third, the comparative analysis highlighted that the customized prompting framework is pivotal in enhancing GPT-4's performance on intrinsic metrics. However, GPT-4-Prompted exhibited notable limitations in specific dimensions, particularly in dimensions such as "Direct", "Warn", "Raise Concern (without permission)", and "Structure", where it underperformed compared to both the baseline GPT-4 and human therapists. Addressing these deficits in intrinsic metrics is critical to improving GPT-4-Prompted's performance on extrinsic metrics.

Table IX

COMPARATIVE ANALYSIS OF MI PERFORMANCE: EVALUATING GPT-4-PROMPTED, GPT-4, AND HUMAN THERAPISTS ON INTRINSIC AND EXTRINSIC METRICS[1]

| Categories | Categories of Intrinsic Metric | Intrinsic Metric | Relation Between Features and the UPQ | Cohen's d (GPT-4-Prompted-Human Therapists) | Cohen's d (GPT-4-Prompted-GPT-4) | Importance Scores |
|---|---|---|---|---|---|---|
| MIIN | Advice (without permission) | advice | Negative | -0.478*** | -0.258*** | 5.540897 |
| | Direct | need | Negative | 0.376*** | 0.257*** | 1.457068 |
| | Warn | Physical | Negative | 0.190*** | 0.190*** | 0.182461 |
| | | health | Negative | 0.000 | 0.216*** | 0.605685 |
| | | substances | Negative | -0.138*** | -0.055 | 0.434961 |
| | Raise Concern (without permission) | discrepancy | Negative | 0.723*** | 0.094* | 0.39475 |
| | Confrontation | conflict | Negative | 0.035 | -0.043 | 0.484878 |
| MICO | Reflection | RES (simple reflection) | Positive | -0.133*** | 0.093* | 1.700861 |
| | | REC (complex reflection) | Positive | 0.039 | -0.143*** | 1.930714 |
| Other | Empathy | Perception | Positive | 0.559*** | 0.146*** | 0.7562 |
| | | Comma | Negative | 0.451*** | 1.734*** | 1.14083 |
| | | work | Positive | -0.335*** | 0.136*** | 0.502621 |
| | Professionalism | Apostrophes | Negative | 0.201*** | -0.732*** | 1.710157 |
| | Negotiation | LSM | Negative | 0.367*** | -0.145*** | 0.35326 |
| | Structure | time | Negative | 0.163*** | 0.158*** | 0.261476 |
| | Analytical Thinking | impersonal pronouns | Positive | -0.479*** | -0.444*** | 1.685375 |
| | | article | Negative | 0.406*** | -0.147*** | 1.746483 |

## V. DISCUSSION

This study utilized a dataset containing nearly 600 real MI dialogues. First, based on an integrative modeling framework, machine learning, deep learning, and NLP techniques were employed to predict human responses, thereby developing a predictive model to assess the UPQ of the LLM within the context of MI. Next, grounded in the MI theory model proposed by Miller and Rollnick, explanatory modeling was conducted to identify key MI strategies and psychological linguistic cues employed by therapists that influence UPQ. Following this, building on the outcomes of the explanatory modeling, we proposed a computational evaluation framework that integrates both extrinsic and intrinsic metrics. Additionally, leveraging this computational evaluation framework, we tailored CoT prompts for the LLM and collected the responses enhanced by the customized prompts. Subsequently, a comparison of MI outcomes between the LLM, human therapists, and the original unprompted LLM was conducted to validate the effectiveness of the computational evaluation framework in enhancing GPT-4's response quality within the context of MI.

### *A. Principal Findings*

The experimental results indicated that the LLM demonstrated exceptional MI performance, with significant improvements observed when customized prompts derived from the comprehensive computational evaluation framework were applied. Overall, the methodologies employed in this study effectively enhanced the LLM's MI performance while ensuring transparency and interpretability of the optimization process. The comprehensive computational evaluation framework combined the extrinsic metric with intrinsic metrics, revealing the intricate relationship between therapist skills and interview outcomes. This approach provided new analytical tools and perspectives for a deeper understanding and optimization of MI practices.

*1) GPT-4's Capabilities in MIs Compared to Human Experts*

LLMs hold significant potential to enhance access to mental health support through scalable interventions capable of reaching extensive populations[34], [22]. To illustrate this potential, developers and end-users have shared anecdotal evidence on social media and other platforms, suggesting that LLMs, such as ChatGPT, exhibit remarkable attributes akin to human therapists, including therapeutic alliance and active listening skills[21], [104]. Consequently, there has been growing consideration among both developers and users to deploy these models as alternatives to human therapists and established, evidence-based psychotherapeutic modalities. As a result, an increasing number of individuals are turning to LLM-based interventions to address their mental health concerns[19], [21], [104]. However, our research indicates that LLMs, such as GPT-4, perform significantly below human therapists in terms of the UPQ, particularly in facilitating high-quality MI interactions. Human therapists scored substantially higher on the UPQ metric (52.69% versus 38.45%, $\chi^2 = 25.886$, $p < 0.001$), highlighting the persistent limitations of LLMs in managing complex psychological interactions and eliciting emotional resonance. These results suggest that LLMs still face significant challenges in delivering effective MI, especially when dealing with intricate emotional and interpersonal dynamics. This underscores the need for further research to ensure the provision of high-quality care and to address the ethical and practical challenges associated with integrating LLMs into mental health services. These findings align with expert concerns regarding the lack of robust evidence supporting the efficacy of LLMs and the potential risks linked to their use in mental health support contexts[22], [23], [24]. Ensuring that mental health interventions maintain high standards of care remains paramount, and ongoing studies are essential to validate the role of LLMs in this sensitive and critical field.

---

[1] Asterisks in Table IX denote levels of statistical significance: * for p < 0.05, ** for p < 0.01, and *** for p < 0.001.

*2) The Role of the UPQ-Centered Evaluation Framework in Assessing GPT-4's Performance in the Processes and Outcomes of Mis*

This study categorizes the intrinsic metrics into three primary types—MIIN behaviors, MICO behaviors, and other behaviors[105] —as introduced in the previous sections. MIIN behaviors refer to actions that conflict with the core principles of Motivational Interviewing (MI), thereby exerting a detrimental effect on UPQ. MICO behaviors, in contrast, align with MI principles[38] and help improve UPQ by fostering client engagement and collaboration. Other behaviors are those that do not fit neatly into either category but still influence the therapy process.

This research indicates that certain LLM-driven behaviors function as a "double-edged sword" in psychotherapy applications. On the positive side, GPT-4's emphasis on complex reflections—a critical MICO behavior—helps deepen the exploration of client issues. This aligns closely with high-quality MI sessions[106]. Moreover, the incorporation of perception-related terminology heightens empathy, as it allows therapists to accurately reflect the client's lived experiences, thus enhancing self-reflection and motivation[107], [108], [109], [110]. On the negative side, however, GPT-4 may also over-utilize behaviors characteristic of low-quality therapy. These include excessive use of directive vocabulary ("have to", "need", "must"), which can be perceived as authoritarian (Direct) and raise client distress when combined with terms like "discrepancy." Such MIIN behaviors undermine UPQ by increasing client anxiety. In addition, overusing apostrophes can lead to an overly informal tone[111], while excessive use of definite articles[112] and LSM[113] may shift the focus away from rapport-building or negotiation tasks. These actions deviate from recommended therapeutic practices[114], [115] and risk damaging the therapeutic alliance and quality of care.

A moderate reduction in some MIIN behaviors can be advantageous. For instance, fewer "warning" expressions about health or substance use may lower client resistance, while minimizing frequent silent pauses can help sustain empathic engagement[116], [117]. These adjustments can elicit more self-motivated client statements[108]. However, overly curtailing beneficial MICO behaviors—particularly simple reflections—poses a threat to MI efficacy. Simple reflections help therapists grasp the client's perspective and daily challenges[118], [119]. When GPT-4 reduces simple reflections excessively, it risks missing crucial nuances in the client's work or personal life, thereby dampening empathic capacity, and increasing relational resistance. Balancing the frequency of simple versus complex reflections is thus critical to optimize the therapeutic alliance and support client autonomy.

Our findings also reveal that general-purpose LLMs, including GPT-4, often overemphasize problem-solving while underemphasizing crucial MI behaviors such as open-ended questioning. Although problem-solving skills can enhance user satisfaction in general contexts, they may inadvertently replicate low-quality therapy traits. This imbalance possibly stems from Reinforcement Learning from Human Feedback (RLHF) processes, which primarily optimize for short-term or generic user metrics[120], [121]. Within mental health settings, however, successful MI relies on long-term engagement and sustained behavioral changes[122]. While GPT-4 may reduce comma usage or utilize more perception-related expressions—both beneficial for MI—excessive reliance on contractions (apostrophes), articles, and demand-related vocabulary like "need" could undermine client autonomy and rapport. To accommodate diverse human preferences and domain-specific requirements, a multi-stage RLHF alignment is necessary, considering not only immediate "usefulness" but also adherence to psychotherapeutic guidelines[120], [121]. Recent studies support the idea of multi-value alignment[123], [124], underscoring the importance of integrating human-centered approaches for healthcare scenarios.

*3) The Facilitating Role of the Integrative computational Evaluation Framework in Prompting LLM Performance in Mis*

As LLMs become increasingly integrated into both novel and existing mental health interventions—spanning commercial sectors[12], [14] and academic environments[13], [15], [125]—establishing transparent and reliable evaluation methodologies is critically important. The framework proposed in this study offers a promising preliminary exploration toward achieving this objective. Based on the UPQ-centered evaluation, it is evident that GPT-4 can further optimize its Simple Reflections, empathic expressions, and professional communication style to align more closely with core MI principles (e.g., avoiding unsolicited advice and excessive emphasis on client needs). Strengthening these aspects will help maintain client autonomy while reducing relational distance, thereby fostering a more collaborative therapeutic alliance.

In terms of MICO (Motivational Interviewing Consistent) behaviors, GPT-4 with customized prompts increased the use of Simple Reflection—one of the core OARS (Open-ended questions, Affirmations, Reflections, and Summaries) techniques—thereby enhancing its ability to understand and respond to clients' core issues. The model also maintained a high frequency of Complex Reflection, reflecting deeper exploration of the client's inner experiences. Additionally, it excelled in Empathy-related word usage (e.g., increased Perception vocabulary), signifying improved attunement to clients' emotional and cognitive states. Concerning professionalism and analytical thinking, GPT-4 with customized prompts reduced the use of apostrophes (thus less informal) and articles (lowering excessive analytical focus), both of which contributed to more supportive and natural clinician–client interactions.

However, GPT-4 with customized prompts still exhibits shortcomings in certain MIIN behaviors—notably Direct commands, Warn, and Raise Concern (without permission)—where its performance remains below that of human therapists. These behaviors can undermine client autonomy or increase resistance, thereby reducing UPQ. Furthermore, the Structure dimension shows room for optimization, suggesting that GPT-4 may benefit from more refined session organization and smoother transitions in line with MI guidelines. By targeting these gaps, future iterations of GPT-4 can strengthen its alignment with established MI principles and better replicate the nuanced approaches used by experienced human therapists.

Overall, this evaluation framework—emphasizing human–AI collaborative design and customized zero-shot prompts—demonstrates that LLMs can significantly narrow the gap with human therapists, both on extrinsic (UPQ) and intrinsic (MIIN/MICO/other behaviors) metrics. By integrating modeling insights and prompt engineering with feedback from mental health experts[13], [126], developers can ensure that these systems adhere to core psychotherapy considerations of quality, safety, and ethics[127], [128]. This approach not only refines GPT-4's performance but also underscores the critical importance of long-term and domain-specific alignment in mental health applications—thereby maintaining clarity, precision, and client well-being in academic and clinical contexts.

### B. Limitations and Future Research

Despite the progress made, several limitations remain, which highlight opportunities for future research. First, key intrinsic metrics critical to UPQ were systematically identified based on a UPQ-centered computational evaluation framework. This enabled targeted enhancements in the LLM's MI performance on both extrinsic and intrinsic metrics through customized zero-shot prompts. Despite these advancements, the LLM still exhibited notable deficiencies in reflective listening and empathetic expression, particularly in complex emotional contexts. These limitations hindered its ability to build emotional resonance, foster trust, and drive self-awareness and behavioral change, ultimately constraining the depth and effectiveness of MI interactions. Comparative analysis further revealed that an excessive reliance on informal language diminished the perceived professionalism of the LLM, whereas an overemphasis on analytical thinking exacerbated emotional detachment from clients. To mitigate these issues, the LLM should limit the use of apostrophes to enhance its professionalism, while increasing the use of impersonal pronouns or reducing articles to mitigate analytical thinking, fostering stronger relational connections with clients. Additionally, although customized prompts have significantly improved LLM's performance in MI, there are still gaps in its MIIN behavior and structural dimensions. In terms of MIIN behavior, the LLM tended to employ direct commands, warnings, and unsolicited expressions of concern, which contravened fundamental MI principles of collaboration and client autonomy. Future research should focus on advancing prompt engineering techniques and refining model fine-tuning to address these deficiencies and further optimize the LLM's ability to adhere to MI principles.

Second, this study employed machine learning-based automated evaluation methods which, while effective in reducing evaluator biases and minimizing cultural influences, fell short of comprehensively capturing interviewers' subjective experiences. Although these methods offered a standardized and scalable evaluation framework, they lacked alignment with real-world interactions and often failed to accurately reflect clients' authentic feedback or the MI quality. To overcome these limitations, future research should consider combining automated and manual evaluation approaches, thereby enhancing the robustness of the assessment process and provide a more nuanced and accurate representation of the MI effectiveness.

Third, the dataset utilized in this study was derived from authentic MI dialogues, however, several limitations must be acknowledged. First, sample selection bias may result in an overrepresentation of certain groups, thereby restricting the generalizability of the findings[129]. Second, social desirability bias could lead to distorted responses from participants, potentially compromising the authenticity of the data[1]. Lastly, cultural diversity issues remain a significant challenge, introducing variability in the quality of MI services[130]. Racial and ethnic minorities (REM), along with other cultural minority groups, are disproportionately likely to prematurely terminate therapy[131] and are less likely to access high-quality care[132].

In conclusion, despite customized prompts, LLM remains less effective than human therapists in managing MIIN behaviors due to insufficient training in nuanced therapeutic communication[133]. Enhancing its proficiency will require integrating targeted training data and refining prompt engineering to better align with MI principles[59]. Furthermore, automated evaluation methods lack the capacity to capture real-world complexities, emphasizing the importance of integrating manual approaches to enhance the precision and robustness of MI assessment. Finally, dataset biases and cultural diversity limitations constrain the generalizability of MI findings, underscoring the need for more inclusive sampling and bias-mitigation strategies in future research.

### C. Theoretical and Practical Implications

#### 1) Theoretical Significance

From the theoretical significance, this study developed a computational evaluation framework to identify key differences in verbal behaviors between LLMs and human therapists during MI. This framework validated and extended Miller and Rollnick's MI theoretical model within human-AI contexts, laying the foundation for the development of more comprehensive and high-fidelity LLM behavior evaluation frameworks in psychotherapy. Secondly, since the introduction of Eliza, a simulated psychotherapist[134], in the 1960s, the concept of therapy chatbots offering mental health support has consistently attracted the interest of clinicians, researchers, and the general public. The findings of this study demonstrate that LLMs perform comparably to human therapists in MI, underscoring their groundbreaking capability to simulate human intelligence and engage in meaningful interactions. These discoveries broaden the scope of AI-human interaction research and underscore the growing significance of machine behavior and machine psychology within the field.

*2) Practical Significance*

From the practical significance, first, the study demonstrated that LLMs approached the MI performance of human therapists, highlighting their potential for mental health applications. As global demand for mental health support continues to exceed the availability of professional therapists, LLMs like ChatGPT—offering therapeutic capabilities, efficiency, cost-effectiveness, and accessibility—emerge as promising tools to help bridge the gap between the supply and demand for mental health services.

Second, a computational evaluation framework developed, applied, and validated in this study offers an effective means for rapidly and objectively assessing AI performance in MI. This framework can assist regulatory bodies and professional organizations in establishing relevant standards and guidelines to ensure the quality and safety of LLM applications. Furthermore, it demonstrates how human expertise has been explicitly integrated to guide LLM behavior through human-AI collaboration. As a transparent and publicly available tool, this framework plays a critical role in guiding and enhancing the capabilities of LLMs, offering new insights into their responsible and ethical deployment in sensitive domains such as psychotherapy.

Furthermore, the implementation of customized prompts offers practical guidance for the real-world deployment of AI systems. These prompts can enhance the professionalism and effectiveness of LLMs, ensuring alignment with MI core principles while improving the quality of interactions with clients.

Finally, this study pioneers the integration of prompt engineering within an integrative modeling paradigm, validating the potential of combining predictive modeling, explanatory modeling, and field experiments in a computational social science research framework. This comprehensive approach enables precise prediction, in-depth understanding, and strategic intervention of LLM language behaviors, thereby advancing the emerging field of human-LLM interaction behavior studies.

### D. Ethical Risks and Potential Biases

Before applying LLMs in the field of MI, it is critical to address several ethical considerations, including data privacy, algorithmic bias, discrimination, transparency, interpretability, and the absence of a comprehensive ethical and legal framework[135]. These ethical risks and inherent biases are particularly pronounced when evaluating the comparative capabilities of human therapists and LLMs in MI.

First, data privacy is a paramount concern, as it involves sensitive personal emotions, psychological states, and private client information. When leveraging LLMs for MI, it is crucial to ensure that real-world data collection and processing adhere to privacy regulations and are safeguarded throughout the processsss[136]. Improper data handling could lead to the leakage of sensitive information, thereby violating clients' privacy rights.

Second, algorithmic bias and discrimination are key ethical considerations. The training data used for LLM may contain inherent biases, which could be amplified during the model's analysis and generation process[137], [138], [139]. In the context of MI, such biases could result in misunderstandings or unfair treatment of specific groups, potentially compromising the fairness and effectiveness of the counseling process. In contrast, while human therapists may also exhibit subjective biases, their professional training equips them to mitigate these biases through self-reflection and supervision.

Additionally, transparency and interpretability are crucial for building trust among stakeholders[138]. Although this study has made some progress in improving transparency and interpretability, several key ethical challenges remain unresolved. For instance, the decision-making process of LLMs is inherently complex and opaque, making it challenging for both clients and regulators to understand the rationale behind the advice provided. In contrast, human therapists can clearly articulate their counseling methods and thought processes, fostering greater understanding and trust.

Finally, the absence of a comprehensive ethical and legal framework poses a significant challenge. As generative AI technologies evolve at a rapid pace, existing ethical and legal norms may fail to keep up with their applications[135]. This misalignment could create regulatory gaps, hindering efforts to effectively protect clients' rights. In contrast, human therapists operate under well-defined professional ethical guidelines and legal regulations, which provide explicit standards for their conduct.

## VI. CONCLUSION

Overall, this study, based on the integrated modeling methodology, combined predictive and explanatory modeling results with the MI theoretical model to develop a UPQ-centered evaluation framework. Theis framework bridges extrinsic metrics and intrinsic metrics, uncovering key intrinsic factors that influence extrinsic outcomes. Leveraging the framework's insights, customized prompts were designed to enhance GPT-4's performance in motivational interviewing. A comparative analysis revealed that while GPT-4's overall performance on intrinsic and extrinsic metrics slightly lagged behind that of human therapists, its MICO behaviors were comparable. Notably, with the implementation of customized prompts, GPT-4 demonstrated significant improvements across all metrics, approaching the performance levels of human therapists. This study offers preliminary evidence of LLMs' capabilities in MI. The findings suggest a cautiously optimistic outlook for applying advanced LLMs in MI interventions, though persistent challenges, particularly in analyzing complex emotions and expressing empathy. The proposed framework emphasizes the integration of human expertise into LLMs top enhance their effectiveness and support the development of trustworthy AI-based MI services. These findings provide direction for further optimizing the application of LLMs in MI and offer important theoretical and practical insights for future research.